\pdfoutput=1

\documentclass[11pt]{article}

\usepackage[]{EMNLP2023}

\usepackage{times}
\usepackage{latexsym}

\usepackage[T1]{fontenc}

\usepackage[utf8]{inputenc}

\usepackage{microtype}

\usepackage{inconsolata}

\usepackage{kotex}
\usepackage{graphicx}
\usepackage{booktabs}
\usepackage{multirow}
\usepackage{amsmath}
\usepackage{array}
\usepackage{listings}
\usepackage[utf8]{inputenc}
\usepackage[T1]{fontenc}
\usepackage{CJKutf8}
\usepackage{lipsum}
\usepackage{ulem}

\title{Efficient Technical Term Translation: A Knowledge Distillation Approach for Parenthetical Terminology Translation}

\author{
  Jiyoon Myung\textsuperscript{*},\hspace{0.5em} Jihyeon Park\textsuperscript{*},\hspace{0.5em} Jungki Son\textsuperscript{*},\hspace{0.5em} Kyungro Lee\textsuperscript{*},\hspace{0.5em} Joohyung Han\textsuperscript{*} \\
  PrompTart LAB, MODULABS \\
  \texttt{\{jiyoon0424, milhaud1201, aeolian83, lkr981147, ddang8jh\}@gmail.com} \\
\small{\textsuperscript{*} All authors contributed equally to this work.}
}

\begin{document}
\maketitle
\begin{abstract}
This paper addresses the challenge of accurately translating technical terms, which are crucial for clear communication in specialized fields. We introduce the Parenthetical Terminology Translation (PTT) task, designed to mitigate potential inaccuracies by displaying the original term in parentheses alongside its translation. To implement this approach, we generated a representative PTT dataset using a collaborative approach with large language models and applied knowledge distillation to fine-tune traditional Neural Machine Translation (NMT) models and small-sized Large Language Models (sLMs). Additionally, we developed a novel evaluation metric to assess both overall translation accuracy and the correct parenthetical presentation of terms. Our findings indicate that sLMs did not consistently outperform NMT models, with fine-tuning proving more effective than few-shot prompting, particularly in models with continued pre-training in the target language. These insights contribute to the advancement of more reliable terminology translation methodologies.
\end{abstract}

\section{Introduction}

Terminology translation task  is essential for understanding documents rich in technical terms, such as academic papers and technical reports. Traditionally, methods in the task have involved identifying term pairs in the source and target languages and using these pairs for training or post-editing purposes. However, challenges arise when there is no precise match for a term in the target language, or when new terms are used inconsistently. For instance, the term "fine-tuning" may be variably translated as "파인튜닝" or "미세조정" in Korean.

To address this, our research proposes a novel approach called Parenthetical Terminology Translation (PTT), which displays the original term in parentheses alongside its translation. This approach aims to mitigate reader confusion, especially when suitable translations are unavailable or translation accuracy is low. Although similar translation strategies using parenthetical form have been suggested in previous studies, effective technical solutions for this approach remain underexplored.

With the advent of advanced Large Language Models (LLMs), researchers have started exploring their potential for various tasks, including translation. LLMs can effectively support PTT through simple prompt usage, offering a promising solution for this approach. However, the practical application of LLMs is hindered by their high computational costs and latency, making them less feasible for real-time or large-scale deployment.

To mitigate these limitations, this study focuses on achieving the capabilities of LLMs using smaller, traditional Neural Machine Translation (NMT) models and small-sized Language Models (sLMs). We generated a high-quality PTT dataset using LLMs and distilled the knowledge by fine-tuning these smaller models with this dataset. This approach ensures that the benefits of LLMs can be harnessed without incurring high computational costs.
Additionally, we evaluated the performance of various models and training methods to optimize model performance and efficiency, particularly for the specialized PTT task.

Our proposed task extends beyond mere translation accuracy; it also emphasizes the correct presentation of technical terms within parentheses, which is crucial for enhancing reader comprehension. To quantitatively evaluate this aspect, we introduced a novel metric specifically designed to assess the models' ability to accurately and effectively use parenthetical annotations. This metric not only evaluates translation quality but also ensures that technical terms are correctly presented, allowing for a robust comparison of model performance across different architectures and training techniques.

Thus, this paper makes three significant contributions to the field of terminology translation:
\begin{enumerate}
  \item \textbf{Synthetic Data Generation}: We propose a collaborative framework using Large Language Models (LLMs) to generate well-curated datasets specifically for the English-Korean Parenthetical Terminology Translation (PTT) task. This framework employs multiple agents to create high-quality sentence pairs, enabling smaller models to perform the PTT task with high accuracy. By leveraging robust data from LLMs, the framework ensures consistency and precision, making it effective for handling domain-specific terminology.
  \item \textbf{Knowledge Distillation and Model Comparison}: Utilizing these high-quality datasets, we fine-tuned various Neural Machine Translation (NMT) models and small-sized Language Models (sLMs). We then conduct a comprehensive performance analysis from diverse perspectives, highlighting the strengths and limitations of each model. This analysis provides valuable insights for future research and development in the field.
  \item \textbf{New Evaluation Metric}: We introduce a novel evaluation metric that quantitatively assesses the ability of models to present appropriate terms within parentheses. This metric ensures contextual accuracy and reader comprehension, offering a robust framework for evaluating model performance in the context of PTT.
\end{enumerate}

These contributions aim to advance the domain of terminology translation by providing practical and efficient solutions that leverage the strengths of both large and small language models. Our approach addresses the inherent challenges of the terminology translation task and paves the way for more accessible translation methodologies in technical and specialized fields.

\section{Related Work}
Terminology translation plays a crucial role in ensuring consistency and accuracy in specialized domains like technical and academic documentation. Early approaches, such as rule-based and statistical machine translation, effectively leveraged pre-defined glossaries and translation memories \cite{Melby1999LEVERAGINGTD}. While these methods successfully maintain consistency within certain contexts, they often struggle with out-of-domain (OOD) words and ambiguous terms \cite{och-ney-2003-systematic}. Moreover, these approaches are less effective when dealing with domain-specific or emerging terms not covered by existing resources \cite{tiedemann-2010-context, tiedemann-scherrer-2017-neural}.

To maintain clarity and precision in academic and technical documentation, it is often necessary to preserve certain terms from the source language. This practice is particularly valuable in cases where the translated term may be unfamiliar to the reader or where retaining the original term is essential for legal or scientific accuracy \cite{moghadam_law_terms, hasler-etal-2018-neural, Michon2020IntegratingDT}. A further strategy to support this practice involves the strategic use of parentheses, where textual additions can help enhance translation quality and consistency through corpus-based improvements \cite{lin-etal-2008-mining, huang-etal-2017-learning-parenthetical, Hawamdeh2018ExplicitationBT}. Despite its potential benefits, the systematic implementation of this approach remains relatively underexplored in current research.

Recent studies have highlighted the effectiveness of knowledge distillation in transferring knowledge from large language models (LLMs) to smaller traditional translation models \cite{li2024mtpatcherselectiveextendableknowledge, enis2024llmnmtadvancinglowresource}. Through this process, datasets generated by a powerful teacher model are distilled into a student model, enabling small-sized models to perform specialized tasks like terminology translation. The multi-agent framework is particularly effective in generating targeted, high-quality datasets for specific tasks \cite{wu2023autogenenablingnextgenllm}. Within this framework, different agents are assigned specialized roles, such as data generation and evaluation, collectively enhancing the quality and relevance of the resulting dataset. This collaborative process is essential for precise and context-aware data generation, which is crucial for training models to excel in specialized translation tasks.

\begin{figure*}[h!]
    \centering
    \includegraphics[width=\textwidth]{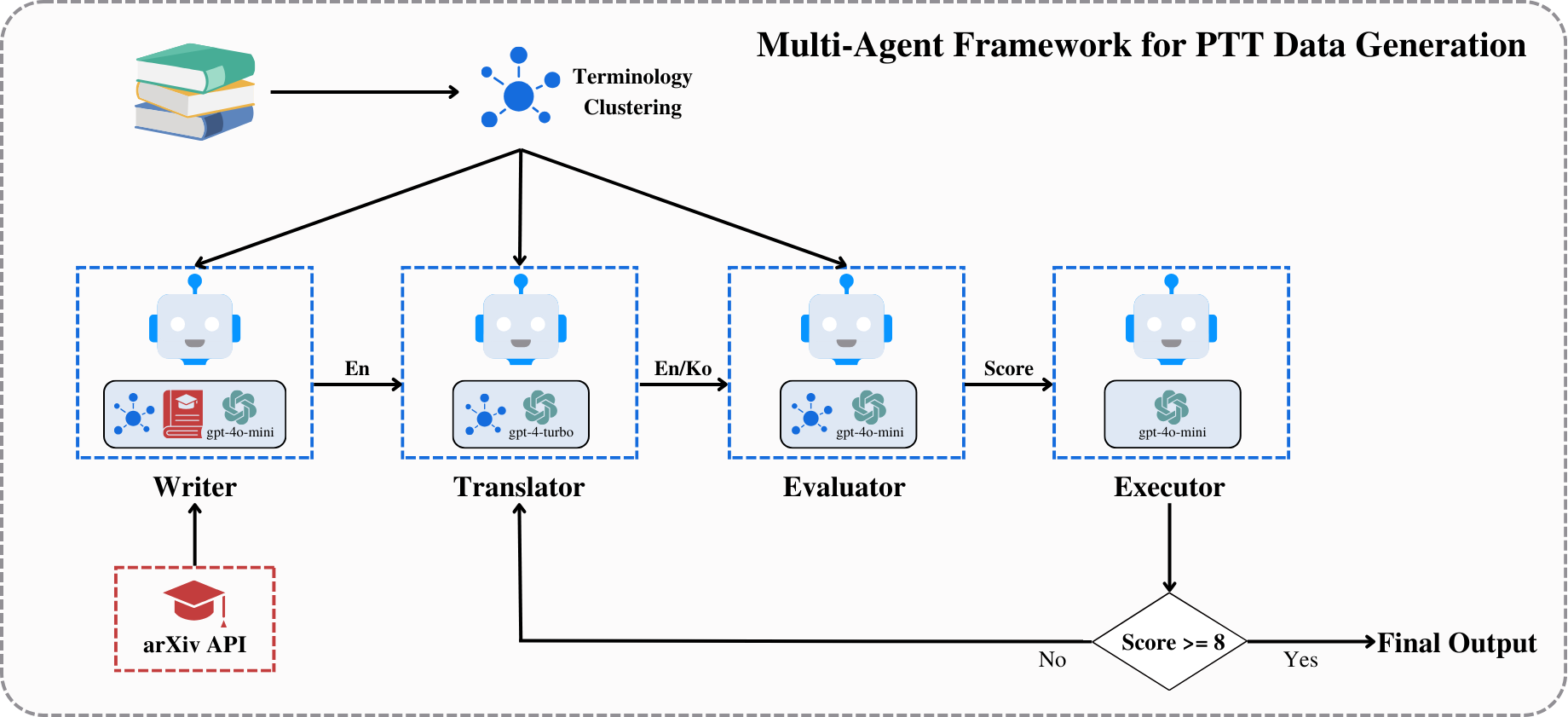}
    \caption{Multi-Agent Framework for generating a high-quality PTT dataset using four agents.}
    \label{fig:multi_agent_framework}
\end{figure*}

\section{Data Generation}

To create a high-quality dataset for the Parenthetical Terminology Translation (PTT) task, we employed four collaborative agents—Writer, Translator, Evaluator, and Executor—utilizing the large language models (LLMs) GPT-4o-mini or GPT-4-turbo for each agent. Our goal was to generate English sentences containing technical terminologies alongside their Korean translations, with the original English terms included within parentheses. The overall framework is illustrated in Figure \ref{fig:multi_agent_framework}.

\subsection{Writer}

The Writer agent was responsible for creating academic English sentences that included the technical terms. To achieve this, we first compiled a comprehensive list of terms to be included in the dataset. Recognizing the rapid emergence of new terminology in the field of artificial intelligence (AI), we focused on terms frequently encountered in AI-related research. To ensure multiple terms could be incorporated into single sentences, we clustered similar domain-specific terms together.

Next, we utilized the arXiv API to find papers that contained all or some of the terms from each cluster. By including the summary of the most relevant paper in the prompt, we helped the Writer LLM understand the appropriate contexts in which these terms were used. This ensured that the generated sentences were contextually accurate and meaningful.

To enhance data diversity, the Writer were tasked with generating sentences where each term appeared either once or in conjunction with other terms.  By combining these sentences post-generation, we facilitated the creation of sentences with various characteristics: sentences where terms appear only once, sentences containing different terms together, and sentences where the same term appears more than twice. This diversity allowed us to analyze the performance of PTT from multiple perspectives, ensuring a comprehensive evaluation of the models under different conditions.
The complete prompt used for the Writer agent is provided in the Appendix (see Listing \ref{lst:writer}).

\subsection{Translator}

The Translator agent translated the English sentences into Korean, ensuring that each target term was followed by its original English term in parentheses to fulfill the PTT task requirements. To enhance accuracy, we employed the GPT-4 Turbo model, while other agents utilized GPT-4o-mini. Additionally, we applied one-shot prompting by providing a relevant example to guide the translation process. This approach helped maintain consistency and precision in the PTT task, ensuring that technical terms were accurately presented within parentheses.
The complete prompt used for the Translator agent is provided in the Appendix (see Listing \ref{lst:translator}).

\begin{table*}[h!]
\centering
\fontsize{8.5pt}{13pt}\selectfont 
\begin{tabular}{l|p{13cm}}
\toprule
\textbf{Terms Set Index} & 0\\
\midrule
\textbf{Domain} & AI (in-domain)\\
\midrule
\textbf{Terms} & adversarial training, recurrent neural architectures, bayesian optimization\\
\midrule
\textbf{Source} & By implementing \textcolor{red}{adversarial training}, researchers have demonstrated significant improvements in the performance of \textcolor{red}{recurrent neural architectures} against adversarial attacks. The integration of \textcolor{red}{recurrent neural architectures} with \textcolor{red}{Bayesian optimization} enhances the model's ability to learn from limited data while minimizing computational resources. \\
\midrule
\textbf{Target} & \textcolor{blue}{적대적 훈련(adversarial training)} 도입으로 연구자들은 적대적 공격에 대응하는 \textcolor{blue}{순환 신경 구조(recurrent neural architectures)}의 성능 향상을 크게 입증하였습니다. \textcolor{blue}{순환 신경 구조(recurrent neural architectures)}와 \textcolor{blue}{베이지안 최적화(bayesian optimization)}를 결합함으로써 모델이 제한된 데이터로부터 학습하는 능력을 향상시키고, 계산적 자원 사용을 최소화합니다. \\
\bottomrule
\end{tabular}
\caption{Sample of generated data from the AI domain (in-domain). Each entry includes the term set index, domain, source text, and the corresponding target translation. Red text highlights the targeted terms \( T_{\text{Eng}} \), while blue text indicates the correct representation of terms \( T_{\text{Kor}} \) in the Korean translation.}
\label{tab:sample_data}
\end{table*}

\subsection{Evaluator/Executor}

The Evaluator agent reviewed the translated sentences, scoring them from 0 to 10 based on the accuracy of term usage and overall translation quality. Then, The Executor agent transit the statement `If the score is less than 8: Response "translator". If the score is 8 or greater: Response "final output".` If a sentence scored below 8, the Evaluator suggested corrections, prompting the Translator to revise the translation. The Translator would then repeat the translation task until the Executor rated the sentence as `"final output"`, ensuring the highest quality and consistency in the dataset.
The complete prompt used for the Evaluator agent is provided in the Appendix (see Listing \ref{lst:evaluator}).

After the automatic data generation process, human reviewers conducted a final quality check to ensure the dataset's reliability. Following this review, we combined seven sentences from each cluster into three composite sentences as mentioned earlier, resulting in 1,398 English-Korean paired sentences encompassing 233 term clusters (a total of 699 distinct terms).  We split these 1,398 sentences into 1,116 for training, 144 for validation, and 138 for testing the performance. We carefully ensured that sentences containing the same terms were allocated to the same dataset, maintaining consistency and preventing data leakage across the splits. The sample data is provided in Table \ref{tab:sample_data}. The entire dataset is available on Hugging Face at \url{https://huggingface.co/datasets/PrompTart/PTT_en_ko}.

\subsection{Out-of-Domain Evaluation Dataset}

To evaluate the generalization ability of models in the PTT task, we generated additional datasets in domains beyond artificial intelligence (AI), specifically targeting biology and physics. The data generation process followed the same methodology as the in-domain training dataset to ensure consistency. In total, we generated 171 paired sentences for biology (subcellular processes), 60 for nanoscale physics, and 168 for high-energy physics. Each domain-specific dataset was curated by referencing relevant academic papers, providing authentic and contextually accurate examples. These out-of-domain datasets allowed for a comprehensive assessment of the models' robustness and adaptability across different specialized fields.

\section{Knowledge Distillation}
In this study, we applied knowledge distillation to fine-tune both traditional neural machine translation (NMT) models and small-sized large language models (sLMs) using the synthetic Parenthetical Terminology Translation (PTT) dataset generated in the previous step. Our goal was to evaluate the effectiveness of distillation techniques across various model architectures, sizes, and training methodologies, offering insights into how distilled models perform in specialized translation tasks.

\subsection{Fine-tuning Traditional Machine Translation Models}

To evaluate the performance of knowledge distillation on traditional neural machine translation models, we employed several widely used open-source models. We focused on encoder-decoder Transformer-based models that support Korean. Specifically, we tested the following models:
\begin{itemize}
 \item mBART50 \cite{liu-etal-2020-multilingual-denoising} : This multilingual NMT model is pre-trained on monolingual corpora from 50 languages and is fine-tuned for translation tasks. It consists of 611 million parameters.
 \item M2M100 \cite{fan2020englishcentricmultilingualmachinetranslation}: A large-scale multilingual NMT model trained on 2200 translation directions, enabling many-to-many translation across 100 languages. We tested the base version with 418 million parameters. 
 \item NLLB-200 \cite{koishekenov-etal-2023-memory}: Known for its extensive language coverage, this model is particularly useful for low-resource languages and inclusive translation services. We tested the distilled version with 600 million parameters.
\end{itemize}
The fine-tuning parameters were provided in the Appendix (see Table \ref{tab:hyperparameters}).

\subsection{Fine-tuning small-sized Large Language Models}

To effectively compare performance with NMT models, we also fine-tuned open-source small-sized large language models. Our goal was to evaluate various models from multiple perspectives to gain comprehensive insights into the PTT task. To ensure reproducibility and a broad evaluation, we selected four well-known and high-performing open-source LLMs:
\begin{itemize}
 \item Llama 3 \cite{touvron2023llamaopenefficientfoundation}: The latest iteration of the Llama series, this model further refines the architecture introduced in earlier versions, enhancing its performance on large-scale datasets. We evaluated the 8B and 70B versions in our experiments.
 \item Gemma 2 \cite{gemmateam2024gemma2improvingopen}: A next-generation multilingual model, Gemma 2 is designed to deliver high performance across diverse natural language tasks with an emphasis on efficiency. We assess the model by testing three versions: the smallest (2B), a mid-sized variant (9B), and a larger configuration (27B).
 \item Qwen 2  \cite{yang2024qwen2technicalreport}: An updated version of the Qwen series, Qwen 2 is developed with a strong focus on flexibility and adaptability to domain-specific tasks. It offers improved performance and efficiency, particularly in handling complex language modeling challenges. In this study, we analyzed the 1.5B, 7B, and 72B versions.
 \item Mistral \cite{jiang2023mistral7b}: Mistral is known for its streamlined design and high efficiency in multilingual tasks. We specifically evaluate the 7B version to examine how its architecture balances performance with computational efficiency.
\end{itemize}
To compare pre-trained models with those that have been further fine-tuned specifically for the Korean language, we also tested models that underwent continual pre-training \cite{ke2023continualpretraininglanguagemodels} in Korean. This approach allowed us to assess the impact of additional language-specific pre-training on the models' performance in the Parenthetical Terminology Translation (PTT) task.
\begin{itemize}
 \item beomi/Llama-3-Open-Ko-8B\href{https://huggingface.co/beomi/Llama-3-Open-Ko-8B}{\footnote{\url{https://huggingface.co/beomi/Llama-3-Open-Ko-8B}}}: A specialized version of Llama 3 focused on Korean language tasks. This open-source model is fine-tuned to excel in Korean linguistic applications. 
  \item beomi/Llama-3-KoEn-8B\href{https://huggingface.co/beomi/Llama-3-KoEn-8B}{\footnote{\url{https://huggingface.co/beomi/Llama-3-KoEn-8B}}}:  A bilingual version of Llama 3 tailored for both Korean and English language tasks. This model is designed to maintain balanced performance across both languages, making it versatile for multilingual applications.
\end{itemize}

Furthermore, we explored instruction-tuned versions of the aforementioned models using different training techniques, such as LoRA \cite{hu2021loralowrankadaptationlarge} and QLoRA \cite{dettmers2023qloraefficientfinetuningquantized}. In addition, we applied few-shot prompting \cite{brown2020languagemodelsfewshotlearners} to both the instruction-tuned models and a commercial LLM (GPT-4o) to compare the effects of fine-tuning versus prompting.
This comprehensive evaluation provides valuable insights into how knowledge distillation, combined with various tuning and prompting strategies, can enhance translation accuracy while maintaining efficiency across diverse model architectures.

The hyper-parameters for fine-tuning and LoRA are detailed in the Appendix (see Table \ref{tab:hyperparameters}), while the full prompt used for few-shot prompting is identical to the Translator agent's prompt (Listing \ref{lst:translator}), with the exception that we did not provide a list of terms in this case.

\section{Custom Evaluation Metric}

This section introduces a novel metric designed specifically for the Parenthetical Terminology Translation (PTT) task, aimed at evaluating not only the accuracy of overall translation but also the correct presentation of the technical terms within parentheses.

For each sentence in the dataset, let \( T_{\text{Eng}} \) represent the list of all technical terms provided in the original English sentence, including duplicates if the same term appears multiple times. Similarly, let \( T_{\text{Kor}} \) represent the list of those terms that are correctly translated into Korean and accompanied by their original English terms in parentheses. We define \( |T_{\text{Eng}}| \) as the total number of technical terms in the English sentence (including duplicates), and \( |T_{\text{Kor}}| \) as the number of correctly translated terms from \( T_{\text{Eng}} \) in the Korean sentence. The ratio of these terms is calculated as the weight \( W_{\text{terms}} = \min\left(\frac{|T_{\text{Kor}}|}{|T_{\text{Eng}}|}, 1\right) \). This ratio is capped at 1 to ensure that no penalty is applied if more terms appear correctly in the Korean translation than in the original English sentence. The adjusted metric for the PTT task, \( M_{\text{PTT}} \), is then computed by multiplying this clipped ratio with the original translation metric \( M \), such that \( M_{\text{PTT}} = W_{\text{terms}} \times M \). Finally, we average \( M_{\text{PTT}} \) across all sentence pairs in the dataset to obtain the final evaluation metric.

We employed BLEU \cite{papineni-etal-2002-bleu}, COMET \cite{rei-etal-2020-comet}, and BERTScore \cite{zhang2020bertscoreevaluatingtextgeneration} as \( M \) to evaluate pure translation performance. The translation metrics are computed after removing the parenthetical terms, ensuring that we assess only the translation's accuracy and fluency. This approach allows us to maintain a focus on both the translation's quality and the correct handling of technical terms within parentheses.

\section{Evaluation}

\begin{table*}[t]
\centering
\fontsize{8.5pt}{9pt}\selectfont 
\begin{tabular}{@{}ccccccc@{}}
\toprule
\textbf{Model} & \textbf{\#Params} & \textbf{Training Techniques} & \textbf{\( W_{\text{terms}} \)} & \textbf{\( M_{\text{PTT}} \) (BLEU)} & \textbf{\( M_{\text{PTT}} \) (COMET)} & \textbf{\( M_{\text{PTT}} \) (BERT)} \\ 
\midrule
\multicolumn{7}{c}{\textbf{Open-source NMT systems}} \\
\midrule
mBART50         & 611M    & Full Fine-Tuning   & 0.931 & 37.519 & 0.831 & 0.863 \\
M2M100         & 418M    & Full Fine-Tuning   & 0.958 & 40.048 & 0.855 & 0.889 \\
NLLB-200       & 600M    & Full Fine-Tuning   & 0.685 & 24.544 & 0.606 & 0.630 \\
\midrule
\multicolumn{7}{c}{\textbf{Llama family sLMs}} \\
\midrule
Llama3         & 8B      & LoRA               & 0.959 & 37.632 & 0.856 & 0.887 \\
               & 8B      & QLoRA              & 0.949 & 38.875 & 0.847 & 0.880 \\
               & 70B     & LoRA               & 0.957 & 38.869 & 0.855 & 0.888 \\
Llama3-Instruct& 8B-it   & QLoRA              & 0.954 & 37.840 & 0.851 & 0.881 \\
               & 8B-it   & 1-shot prompting   & 0.523 & 0.577  & 0.214 & 0.310 \\
\midrule
\multicolumn{7}{c}{\textbf{Gemma family sLMs}} \\
\midrule
Gemma2         & 2B      & LoRA               & 0.946 & 37.959 & 0.842 & 0.875 \\
               & 9B      & LoRA               & 0.958 & 41.567 & 0.858 & 0.893 \\
               & 9B      & QLoRA              & 0.935 & 38.955 & 0.835 & 0.869 \\
               & 27B     & LoRA               & 0.966 & 40.856 & 0.865 & 0.899 \\
Gemma2-Instruct& 9B-it   & QLoRA              & 0.953 & 39.215 & 0.849 & 0.884 \\
               & 9B-it   & 1-shot prompting   & 0.342 & 9.698  & 0.286 & 0.286 \\
\midrule
\multicolumn{7}{c}{\textbf{Qwen family sLMs}} \\
\midrule
Qwen2          & 1.5B    & LoRA               & 0.950 & 34.374 & 0.838 & 0.868 \\
               & 7B      & LoRA               & 0.945 & 39.167 & 0.844 & 0.877 \\
               & 7B      & QLoRA              & 0.951 & 38.014 & 0.846 & 0.879 \\
               & 72B     & LoRA               & 0.956 & 40.837 & 0.855 & 0.889 \\
Qwen2-Instruct & 7B-it   & QLoRA              & 0.947 & 37.990 & 0.842 & 0.874 \\
\midrule
\multicolumn{7}{c}{\textbf{Mistral family sLMs}} \\
\midrule
Mistral          & 7B      & QLoRA              & 0.931 & 37.646 & 0.830 & 0.862 \\
Mistral-Instruct & 7B-it   & QLoRA              & 0.927 & 37.990 & 0.826 & 0.857 \\
\midrule
\multicolumn{7}{c}{\textbf{Korean Continued Pre-trained sLMs}} \\
\midrule
Llama-3-KoEn   & 8B-it   & QLoRA              & \textbf{0.974} & \textbf{41.789} & \textbf{0.873} & \textbf{0.907} \\
               & 8B-it   & 1-shot prompting   & 0.614 & 0.333  & 0.080 & 0.110 \\
Llama-3-Open-Ko & 8B-it  & QLoRA              & 0.953 & 39.869 & 0.852 & 0.885 \\
\midrule
\multicolumn{7}{c}{\textbf{Commercial LLM}} \\
\midrule
GPT-4o         & Unknown & 0-shot prompting   & 0.616 & 20.596 & 0.547 & 0.564 \\
GPT-4o         & Unknown & 1-shot prompting   & 0.751 & 26.509 & 0.669 & 0.689 \\
\bottomrule               
\end{tabular}
\caption{Model performance metrics for in-domain test data. \( W_{\text{terms}} \) represents the average ratio of correctly translated terms with original English terms in parentheses. \( M_{\text{PTT}} \) (BLEU), \( M_{\text{PTT}} \) (COMET), and \( M_{\text{PTT}} \) (BERT) are the original tranlsation metrics adjusted using \( W_{\text{terms}} \) and averaged over all data. The suffix '-it' indicates instruct-tuned models. The top scores for each metric are highlighted in bold.}
\label{tab:result}
\end{table*}

\begin{table*}[t]
\centering
\fontsize{8.5pt}{9pt}\selectfont 
\begin{tabular}{@{}ccccccc@{}}
\toprule
\textbf{Model} & \textbf{\#Params} & \textbf{Training Techniques} & \textbf{\( W_{\text{terms}} \)} & \textbf{\( M_{\text{PTT}} \) (BLEU)} & \textbf{\( M_{\text{PTT}} \) (COMET)} & \textbf{\( M_{\text{PTT}} \) (BERT)} \\ 
\midrule
\multicolumn{7}{c}{\textbf{Open-source NMT systems}} \\
\midrule
mBART50         & 611M    & Full Fine-Tuning   & 0.784 & 19.538 & 0.668 & 0.698 \\
M2M100         & 418M    & Full Fine-Tuning   & 0.763 & 20.472 & 0.650 & 0.680 \\
NLLB-200       & 600M    & Full Fine-Tuning   & 0.160 & 4.066  & 0.138 & 0.144 \\
\midrule
\multicolumn{7}{c}{\textbf{Llama family sLMs}} \\
\midrule
Llama3         & 8B      & LoRA               & 0.769 & 22.595 & 0.670 & 0.693 \\
               & 8B      & QLoRA              & 0.849 & 23.792 & 0.736 & 0.760 \\
               & 70B     & LoRA               & 0.854 & 33.321 & 0.762 & 0.788 \\
Llama3-Instruct& 8B-it   & QLoRA              & 0.864 & 23.498 & 0.750 & 0.775 \\
\midrule
\multicolumn{7}{c}{\textbf{Gemma family sLMs}} \\
\midrule
Gemma2         & 2B      & LoRA               & 0.824 & 24.203 & 0.711 & 0.736 \\
               & 9B      & LoRA               & 0.886 & 38.639 & 0.793 & 0.822 \\
               & 9B      & QLoRA              & \textbf{0.900} & 32.914 & 0.799 & 0.826 \\
               & 27B     & LoRA               & 0.897 & \textbf{40.379} & \textbf{0.804} & \textbf{0.836} \\
Gemma2-Instruct& 9B-it   & QLoRA              & 0.883 & 34.861 & 0.785 & 0.813 \\
\midrule
\multicolumn{7}{c}{\textbf{Qwen family sLMs}} \\
\midrule
Qwen2          & 1.5B    & LoRA               & 0.762 & 9.831  & 0.598 & 0.635 \\
               & 7B      & LoRA               & 0.750 & 18.531 & 0.637 & 0.662 \\
               & 7B      & QLoRA              & 0.849 & 20.028 & 0.729 & 0.751 \\
               & 72B     & LoRA               & 0.877 & 32.048 & 0.779 & 0.806 \\
Qwen2-Instruct & 7B-it   & QLoRA              & 0.864 & 23.498 & 0.750 & 0.775 \\
\midrule
\multicolumn{7}{c}{\textbf{Mistral family sLMs}} \\
\midrule
Mistral           & 7B      & QLoRA              & 0.870 & 15.942 & 0.713 & 0.749 \\
Mistral-Instruct  & 7B-it   & QLoRA              & 0.876 & 17.350 & 0.717 & 0.756 \\
\midrule
\multicolumn{7}{c}{\textbf{Korean Continued Pre-trained sLMs}} \\
\midrule
Llama-3-KoEn    & 8B-it   & QLoRA              & 0.884 & 35.492 & 0.789 & 0.817 \\
Llama-3-Open-Ko & 8B-it  & QLoRA              & 0.887 & 35.409 & 0.790 & 0.813 \\
\bottomrule               
\end{tabular}
\caption{Model performance metrics for out-of-domain test data. \( W_{\text{terms}} \) represents the average ratio of correctly translated terms with original English terms in parentheses. \( M_{\text{PTT}} \) (BLEU), \( M_{\text{PTT}} \) (COMET), and \( M_{\text{PTT}} \) (BERT) are the original tranlsation metrics adjusted using \( W_{\text{terms}} \) and averaged over all data. The suffix '-it' indicates instruct-tuned models. The top scores for each metric are highlighted in bold.}
\label{tab:result_ood}
\end{table*}

\subsection{Quantitative Analysis}
The results presented in Table \ref{tab:result} provide a comprehensive overview of the quantitative performance of various models and training techniques on the in-domain Parenthetical Terminology Translation (PTT) dataset, while Table \ref{tab:result_ood} presents results on the out-of-domain dataset. Key observations are summarized as follows:
\begin{enumerate}
\item \textbf{sLMs vs. NMT Models}: The performance comparison between small-sized Large Language Models (sLMs) and Neural Machine Translation (NMT) models reveals that sLMs do not consistently outperform NMT models, even though LLMs are often perceived as more advanced due to their architecture. For instance, mBART50 and M2M100, achieved weighted BLEU scores of 37.52 and 40.05, respectively, with corresponding weight indicate strong PTT performance. These scores were comparable or superior to those achieved by some sLMs, such as the Llama 3 8B and 70B models, which obtained similar weighted BLEU scores but required significantly larger model sizes.
\item \textbf{Instruction-Tuned vs. Base Models}: Within the same sLM families, base models generally slightly outperformed instruction-tuned models on the PTT task. For instance, the Llama 3 8B model with QLoRA achieved a weighted BLEU score of 38.88, while the instruction-tuned version (8B-it) with the same QLoRA technique scored slightly lower at 37.84. This trend suggests that instruction-tuned models, which are trained on a broad range of tasks, may not gain a specific advantage for the specialized requirements of the PTT task.
\item \textbf{Fine-Tuning vs. Prompt Engineering}: Applying prompt engineering instead of fine-tuning to instruction-tuned models, using a 1-shot prompting approach, resulted in very poor performance. For example, the Llama 3 8B-it scored only 0.523, and the Gemma 2-9B-it scored 0.342 on weight metric. Even the commercial LLM GPT-4o performed worse than other fine-tuned small models, underscoring the critical importance of fine-tuning for specialized tasks like PTT.
\item \textbf{Korean Continued Pre-trained Models}: Models that underwent continued pre-training in the target language (Korean) generally outperformed others, with the Llama-3-KoEn-8B-it achieving the highest score among all models. Although Llama-3-Open-Ko-8B, which was continued pre-trained exclusively in Korean, showed slightly lower performance with a weighted BLEU score of 39.869, it still performed well. This highlights the importance of bilingual proficiency in models for the PTT task, where handling both source and target languages effectively is crucial for success.
\item \textbf{Model Size and Out-of-Domain Performance}: In the in-domain dataset, model size had little impact on performance, with smaller models like Gemma 2 7B even outperforming larger ones like Gemma 2 27B. However, when tested on out-of-domain datasets, all models experienced significant performance drops, but larger models such as Gemma 2 27B or Llama 3 70B showed less decline, indicating better generalization capabilities. This suggests that while smaller models can be highly effective in specialized tasks, larger models are more versatile and better suited for handling diverse and unfamiliar datasets. The larger models' ability to retain higher performance levels in out-of-domain tasks underscores their capacity to adapt to a wider range of terminologies and contexts, making them more versatile in applications where data variability is a key challenge.
\end{enumerate}

\subsection{Qualitative Analysis}

\begin{enumerate}
\item \textbf{Progression of PTT and Translation Skills}: As illustrated in Table \ref{tab:m2m_result_epoch} , most of the models, including M2M100, initially demonstrated strong proficiency in the PTT task, particularly in incorporating original terms within parentheses, as indicated by the high weight metrics in the earlier epochs. Over successive training epochs, the model's overall translation skills improved gradually, leading to better performance across all weighted metrics. A detailed illustration of these improvements is provided in the Appendix (see Table \ref{tab:detailed_output}).

\item \textbf{Challenges with Less Common Terms}: Our analysis highlights a persistent challenge among models in accurately translating less common terms, especially proper nouns. As demonstrated in Table \ref{tab:result_proper_nouns}, terms like "de Finetti's theorem" were inconsistently translated across different models, reflecting the difficulty these models face when dealing with less familiar terminology. This inconsistency underscores the importance of the PTT task, which helps maintain translation accuracy by preserving original terms alongside their translations, thereby reducing the likelihood of incorrect interpretations.

\item \textbf{Out-of-Domain Translation Challenges}: Most models struggled with translating out-of-domain (OOD) sentences, as detailed in Table \ref{tab:deatiled_result_ood}. They often failed to accurately translate OOD terms, frequently substituting them with unrelated or incorrectly adapted words, sometimes even drawing from other languages. These frequent mistranslations highlight the need for more robust training methods or supplementary mechanisms to improve the models' generalization ability for handling unseen datasets effectively.
\end{enumerate}

\begin{table*}[]
\centering
\fontsize{10pt}{13pt}\selectfont 
\begin{tabular}{lcccc}
\toprule
\textbf{Epoch} & \textbf{Weight} & \textbf{Weighted BLEU} & \textbf{Weighted COMET} & \textbf{Weighted BERT Score} \\
\midrule
epoch 1  & 0.939 & 31.780 & 0.824 & 0.858 \\
epoch 3  & 0.924 & 35.668 & 0.821 & 0.853 \\
epoch 5  & 0.948 & 37.853 & 0.844 & 0.878 \\
epoch 7  & 0.956 & 38.685 & 0.851 & 0.886 \\
epoch 9  & 0.958 & 40.048 & 0.855 & 0.889 \\
\bottomrule
\end{tabular}
\caption{Model performance metrics of the M2M100 model for test data across training epochs.}
\label{tab:m2m_result_epoch}
\end{table*}

\section{Conclusion}

In this study, we explored the Parenthetical Terminology Translation (PTT) task, a specialized translation problem that focuses on mitigating potential inaccuracies in term translation by displaying the original technical term in parentheses alongside its translation. To effectively evaluate this approach, we introduced a novel evaluation metric, \( M_{\text{PTT}} \), designed to measure both the accuracy of overall translation and the proper parenthetical presentation, ensuring that technical terms are effectively communicated across languages.

To generate a high-quality dataset for this task, we utilized a collaborative approach involving Writer, Translator, Evaluator, and Executor agents, supported by large language models (GPT-4). This allowed us to create a diverse and contextually accurate dataset that reflects real-world usage of technical terms in artificial intelligence (AI), biology, and physics. We then applied knowledge distillation techniques to fine-tune both traditional Neural Machine Translation (NMT) models and small-sized Large Language Models (sLMs), comparing their performance across various model architectures, sizes, and training methods.

Our findings revealed that sLMs did not consistently outperform NMT models, challenging the assumption that more advanced architectures inherently lead to superior performance. Additionally, within the same sLM families, base models slightly outperformed instruction-tuned models, suggesting that broad task training may not offer advantages for specialized tasks like PTT. Fine-tuning proved crucial, as prompt engineering approaches like 1-shot prompting resulted in significantly poorer performance. Moreover, models with continued pre-training in Korean outperformed others, highlighting the importance of bilingual proficiency for the PTT task. While model size had little impact on in-domain performance, larger models demonstrated better generalization on out-of-domain datasets, suggesting they are more versatile and better suited for handling diverse and unfamiliar data. These insights contribute to optimizing models and training techniques for specialized translation tasks, offering practical guidance for future research and applications in terminology translation.

\section*{Limitations}

\textbf{Penalty Mechanism in Evaluation Metrics}: The current approach to evaluating PTT performance involves simply multiplying translation metrics by a weight that reflects the presence of correctly parenthesized terms. However, this straightforward multiplication can disproportionately affect the overall performance scores. A more sophisticated penalty mechanism, such as using an exponential function, could provide a more balanced assessment by reducing the impact on the metric scores. Additionally, the current metric does not penalize the model for excessively parenthesizing trivial or unintended terms, which could lead to over-parenthesization. Future work could incorporate penalties for such cases, potentially by introducing concepts of recall and precision to refine the evaluation.

\textbf{Potential Bias in the Dataset}: The PTT dataset was generated using GPT-4, and the performance metrics were assessed with this dataset as the ground truth. This approach may introduce biases inherent to the GPT-4 model into the dataset, potentially affecting the robustness and generalizability of the models trained on it. To mitigate this, future research should consider generating datasets using a variety of models, ensuring a broader representation of translation styles and reducing the potential for model-specific biases.

\textbf{Language Scope of the Study}: This study focused exclusively on translation into Korean, which limits the generalizability of the findings across different languages. PTT performance might vary significantly with other languages due to differences in linguistic structures and translation challenges. Expanding the study to include translations into multiple languages would enable a more comprehensive analysis of the PTT task and provide insights into how the models perform across different linguistic contexts.


\section*{Acknowledgements}
This research was supported by Brian Impact Foundation, a non-profit organization dedicated to the advancement of science and technology for all.
\bibliography{anthology}
\bibliographystyle{acl_natbib}
\onecolumn
\newpage
\appendix

\section{Appendix}

\label{sec:appendix}
\begin{table*}[!h]
\centering
\small
\begin{tabular}{l|l|c|c|c}
\toprule
                  & Parameter & NMT      & sLM (w/ LoRA)                & sLM (w/ QLoRA)               \\
\midrule
Training Argument & Learning Rate     & 3e-5     & 1e-4                & 2e-4                \\
& Lr Scheduler Type & linear   & cosine              & cosine              \\
& Optimizer         & AdamW    & paged\_adamw\_32bit & paged\_adamw\_32bit \\
& Warmup Ratio      & N/A      & 0.03                & 0.03                \\
& Weight Decay      & 0.01     & 0.001               & N/A                 \\
& Max Grad Norm     & 1.0      & 1.0                 & 0.3                 \\
& Dtype             & bfloat16 & bfloat16            & bfloat16            \\
\midrule
LoRA Configure & LoRA R            & N/A      & 64                  & 64                  \\
& LoRA Alpha        & N/A      & 16                  & 16                  \\
& LoRA Dropout      & N/A      & 0.1                 & 0.1                 \\
& Target Modules    & N/A      & all-linear          & all-linear         \\
\bottomrule
\end{tabular}
\caption{Hyper-parameters for Fine-Tuning and LoRA Techinque}
\label{tab:hyperparameters}
\end{table*}

\begin{lstlisting}[frame=single, breaklines=true, captionpos=b,
caption={Full Prompt of Writer}, label={lst:writer}, 
backgroundcolor=\color{gray!20}, basicstyle=\ttfamily\footnotesize]

You are a professional paper writer.
    
[TERM1] = {terms[0]}
[TERM2] = {terms[1]}
[TERM3] = {terms[2]}

<reference>
{arxiv_summaries}
</reference>

<instruction>
- The request is to thoroughly review and cite the provided <reference> when writing theacademic paper.
- Write complex English sentences using the given technical terms.
- Use appropriate academic tone.
- Each sentence MUST be clear, accurate, and contextually appropriate for a scientific paper.
- Generate only in English.
</instruction>

## Output Format:
1.english: A sentence using terms [TERM1].
2.english: A sentence using terms [TERM2].
3.english: A sentence using terms [TERM3].
4.english: A sentence using terms [TERM1] and [TERM2].
5.english: A sentence using terms [TERM2] and [TERM3].
6.english: A sentence using terms [TERM1] and [TERM3].
7.english: A sentence using terms [TERM1], [TERM2], and [TERM3].

CAUTION: Ensure that exactly 7 sentences are generated.
\end{lstlisting}

\newpage

\begin{lstlisting}[frame=single, breaklines=true, escapeinside=``, captionpos=b,
caption={Full Prompt of Translator}, label={lst:translator}, 
backgroundcolor=\color{gray!20}, basicstyle=\ttfamily\footnotesize]

You are a professor specializing in AI, proficient in both Korean and English.

[TERM1] = {terms[0]}
[TERM2] = {terms[1]}
[TERM3] = {terms[2]}

<translation guideline>
- Translate while preserving the original term like `사전 훈련`(pre-train).
- If there is an abbreviation, translate it like this Korean term(english term, abbreviation).
- Identify terms, acronyms, and concepts to keep in English.
- Maintain academic tone and technical accuracy in your translations.
- Ensure the translation is natural in Korean while accurately conveying the original meaning.
- Change all the letters within the parentheses in Korean sentences to lowercase.
- IMPORTANT: The terms corresponding to [TERM1], [TERM2], and [TERM3] MUST ALWAYS be enclosed in parentheses like this: Korean term(English term).
</translation guideline>

<example>
english: LLMs demonstrate new abilities such as in-context learning, instruction following, and multi-step reasoning, enabling them to learn new tasks, follow instructions, and effectively solve complex problems.
`\begin{CJK}{UTF8}{mj}
korean: LLM은 맥락 학습(in-context learning), 지시 사항 따르기(instruction following), 다단계 추론
\end{CJK}`
`\begin{CJK}{UTF8}{mj}
 (multi-step reasoning)과 같은 새로운 능력을 보여줌으로써 새로운 작업을 학습하고, 지시를 따르며, 
\end{CJK}`
`\begin{CJK}{UTF8}{mj}
복잡한 문제를 효과적으로 해결할 수 있습니다.
\end{CJK}`
</example>

## Output Format:
1.korean: [Korean translation]
2.korean: [Korean translation]
...
( Continue this pattern for all 7 sentences )

\end{lstlisting}

\newpage

\begin{lstlisting}[frame=single, breaklines=true,  escapeinside=``, captionpos=b,
caption={Full Prompt of Evaluator}, label={lst:evaluator}, 
backgroundcolor=\color{gray!20}, basicstyle=\ttfamily\footnotesize]

You're an expert evaluating English to Korean translations of research papers, with a specific focus on proper parenthetical translations of technical terms.

<criteria>
- The format for parenthetical translations should be: Korean term(English term).
- The specific terms {terms[0]}, {terms[1]} or {terms[2]} MUST ALWAYS be enclosed in parentheses in the Korean translation.
- Parentheses should be properly placed, ensuring consistency in parenthesizing across the entire sentence.
- Ensures the translation conveys the original meaning precisely and reads naturally and smoothly.
</criteria>

<instruction>
- Change all the letters within the parentheses in Korean sentences to lowercase.
- Evaluate the Korean translation of the provided English sentences.
- Check the consistency and correctness of parenthesization.
- Provide a score (0-10) based on the correctness and consistency of parenthesization as Korean term(English term).
- Offer specific improvement suggestions if the score is less than 10.
- DO NOT include any supplementary explanations.
- Check your output format again.
</instruction>

## Example Output:
english: The neural network uses backpropagation to optimize its weights.
`\begin{CJK}{UTF8}{mj}
korean: 신경망(neural network)은 역전파(backpropagation)를 사용하여 가중치(weight)를 최적화합니다.
\end{CJK}`
score: 10/10
terms_check: [neural network: Yes, backpropagation: Yes, weight: Yes]
parentheses_count: 3
suggestions: No improvements needed / Suggest ensuring that "model compression" is 
`\begin{CJK}{UTF8}{mj}
translated as  "모델 압축(model compression)" and adjusting "모델 컴프레션" to "model compression" 
\end{CJK}`
for consistency and clarity.

## Example Format:
1.
english: [English text using term "{terms[0]}"]
korean: [Korean translation using parentheses]
score: [X/10]
terms_check: [{terms[0]}: Yes/No, {terms[1]}: Yes/No, {terms[2]}: Yes/No]
parentheses_count: [Number of parentheses pairs in the Korean translation]
suggestions: [Suggest capturing the original meaning and nuances in the translation while adjusting the structure for natural flow and grammar]
2.
english: [English text using terms "{terms[0]}" and "{terms[1]}"]
korean: [Korean translation]
3.
...
(Continue this pattern for all 7 sentences)

\end{lstlisting}

\begin{table*}[]
\centering
\fontsize{8.5pt}{13pt}\selectfont 
\begin{tabular}{l|p{14cm}}
\toprule
\textbf{Domain} & AI (in-domain) \\
\midrule
\textbf{Terms} & heterogeneous graphs, neural probabilistic text segmentation \\
\midrule
\textbf{Source} & Heterogeneous graphs provide a rich framework for the implementation of neural probabilistic text segmentation, facilitating improved contextual understanding through varied data representations. \\
\midrule
\textbf{Target} & 이종 그래프(heterogeneous graphs)는 신경 확률적 텍스트 분할(neural probabilistic text segmentation)의 구현을 위한 풍부한 프레임워크를 제공하며, 다양한 데이터 표현을 통해 개선된 맥락적 이해를 촉진합니다.\\
\toprule
\textbf{Epoch} & \textbf{Model Output} \\
\midrule
epoch 1 & \textcolor{red}{\sout{헤테로겐}} \textcolor{blue}{이종}  그래프(heterogeneous graphs)는 신경 확률적 텍스트 분할(neural probabilistic text segmentation)의 구현에 풍부한 프레임워크를 제공하며, 다양한 데이터 표현을 통해 맥락적 이해를 향상시킵니다.  \\
\midrule
epoch 2 & \textcolor{red}{\sout{동적}} \textcolor{blue}{이종}  그래프(heterogeneous graphs)는 신경 확률적 텍스트 분할(neural probabilistic text segmentation)의 구현을 위한 풍부한 프레임워크를 제공하며, 다양한 데이터 표현을 통해 맥락 이해를 향상시킵니다.  \\
\midrule
epoch 3 & \textcolor{blue}{이종 그래프(heterogeneous graphs)는} 신경 확률적 텍스트 분할(neural probabilistic text segmentation)의 구현을 위한 풍부한 프레임워크를 제공하며, 다양한 데이터 표현을 통해 문맥 이해를 향상시킵니다. \\
\midrule
epoch 4 & \textcolor{red}{\sout{동성애}} \textcolor{blue}{이종}  그래프(heterogeneous graphs)는 신경 확률적 텍스트 분할(neural probabilistic text segmentation)의 구현을 위한 풍부한 프레임워크를 제공하며, 다양한 데이터 표현을 통해 개선된 맥락 이해를 촉진합니다. \\
\midrule
epoch 5 & \textcolor{red}{\sout{동적}} \textcolor{blue}{이종}  그래프(heterogeneous graphs)는 다양한 데이터 표현을 통해 개선된 맥락 이해를 촉진하여 신경 확률적 텍스트 분할(neural probabilistic text segmentation)의 구현을 위한 풍부한 프레임워크를 제공합니다. \\
\midrule
epoch 6 & \textcolor{red}{\sout{동성애}} \textcolor{blue}{이종}  그래프(heterogeneous graphs)는 신경 확률적 텍스트 분할(neural probabilistic text segmentation)의 구현을 위한 풍부한 프레임워크를 제공하며, 다양한 데이터 표현을 통해 문맥 이해를 개선하는 데 도움을 줍니다.  \\
\midrule
epoch 7 & \textcolor{Green}{이질적인} 그래프(heterogeneous graphs)는 신경 확률적 텍스트 분할(neural probabilistic text segmentation)의 구현을 위한 풍부한 프레임워크를 제공하며, 다양한 데이터 표현을 통해 개선된 맥락 이해를 촉진합니다. \\
\midrule
epoch 8 & \textcolor{Green}{이질적인} 그래프(heterogeneous graphs)는 신경 확률적 텍스트 분할(neural probabilistic text segmentation)의 구현을 위한 풍부한 프레임워크를 제공하며, 다양한 데이터 표현을 통해 개선된 맥락 이해를 촉진합니다. \\
\midrule
epoch 9 & \textcolor{Green}{이질적인} 그래프(heterogeneous graphs)는 신경 확률적 텍스트 분할(neural probabilistic text segmentation)의 구현을 위한 풍부한 프레임워크를 제공하며, 다양한 데이터 표현을 통해 개선된 맥락 이해를 촉진합니다. \\
\bottomrule
\end{tabular}
\caption{Illustration of M2M100 output for in-domain sample data across epochs. Red strikethrough text indicates incorrect translations, with blue text showing the corrected translations. Green text represents correct translations that were maintained or introduced during later epochs.}
\label{tab:detailed_output}
\end{table*}

\begin{table*}[]
\centering
\fontsize{8.5pt}{13pt}\selectfont 
\begin{tabular}{l|p{12cm}}
\toprule
\textbf{Domain} & AI (in-domain) \\
\midrule
\textbf{Terms} & neural task-driven modeling, de Finetti's theorem\\
\midrule
\textbf{Source} & Neural task-driven modeling, when applied to the constructs of de Finetti's theorem, unveils a sophisticated approach to managing uncertainty in predictive models within artificial intelligence. \\
\midrule
\textbf{Target} & 신경 작업 중심 모델링(neural task-driven modeling)이 \textcolor{Green}{드 핀네티의 정리}(de finetti's theorem)의 구조에 적용될 때, 인공 지능 내에서 예측 모델의 불확실성을 관리하는 정교한 접을 드러냅니다. \\
\toprule
\textbf{Model} & \textbf{Model Output} \\
\midrule
mBART50& 신경 작업 주도 모델링(neural task-driven modeling)이 \textcolor{blue}{데 페네티 이론}의 구성 요소에 적용될 때, 인공지능 내 예측 모델에서의 불확실성을 관리하는 정교한 접근 방식을 밝혀냅니다.   \\
\midrule
M2M100 & 신경 작업 주도 모델링(neural task-driven modeling)이 \textcolor{blue}{디 피네티(de Finetti) 이론}의 구조에 적용될 때, 인공 지능 내 예측 모델의 불확실성을 관리하는 정교한 접근 방식을 드러냅니다. \\
\midrule
NLLB-200 & \textcolor{blue}{데 피네티 정리}의 적용은 신경 작업 주도 모델링(neural task-driven modeling)과 함께 사용함으로써 불확실성을 효과적으로 관리할 수 있는 더 탄력적인 모델을 가능하게 합니다. \\
\midrule
\midrule
Llama3-8B & 신경 작업 주도 모델링(neural task-driven modeling)을 \textcolor{blue}{데 피니의 정리}의(de Finetti's theorem) 구축에 적용하면 인공 지능 내 예측 모델의 불확실성을 관리하는 정교한 접근 방식을 드러낼 수 있습니다.  \\
\midrule
Llama3-8B-Instruct & 신경 작업 주도 모델링(neural task-driven modeling)이 \textcolor{blue}{데 피니의 정리}(de finetti's theorem)의 구축에 적용될 때, 인공지능 내에서 예측 모델의 불확실성을 관리하는 정교한 접근 방식을 드러냅니다.  \\
\midrule
Gemma2-9B & 신경 작업 주도 모델링(neural task-driven modeling)이 \textcolor{blue}{디 핀에티의 정}리(de finetti's theorem)의 구조에 적용될 때, 인공지능 내 예측 모델에서 불확실성을 관리하는 정교한 접근 방식을 밝혀냅니다.  \\
\midrule
Gemma2-9B-Instruct & 신경 과제 주도 모델링(neural task-driven modeling)을 \textcolor{blue}{데 핀에티의 정리}(de finetti's theorem)의 구조에 적용하면 인공지능 내 예측 모델에서 불확실성을 관리하는 정교한 접근 방식을 드러냅니다. \\
\midrule
Qwen2-7B & \textcolor{blue}{디-finetti의 정리}(de finetti's theorem)의 구조에 적용된 신경 작업 주도 모델링(neural task-driven modeling)은 인공 지능에서 예측 모델의 불확실성을 관리하는 정교한 접근 방식을 밝혀냅니다.  \\
\midrule
Qwen2-7B-Instruct & \textcolor{blue}{디 베니테이의 정리}(de finetti's theorem)의 구조에 신경 작업 주도 모델링(neural task-driven modeling)을 적용하면 인공 지능에서 예측 모델의 불확실성을 관리하는 정교한 접근 방식을 밝혀낼 수 있습니다.  \\
\midrule
Mistrial-7B & 신경 작업 주도 모델링(neural task-driven modeling)이 \textcolor{blue}{디 피네티(de finetti)의 정리}의 구조에 적용될 때, 인공지능 내에서 예측 모델에서 불확실성을 관리하는 정교한 접근 방식을 밝혀냅니다. \\
\midrule
Mistrial-7B-Instruct & 신경 작업 주도 모델링(neural task-driven modeling)을 \textcolor{blue}{데 핀철리의 정리}(de finetti's theorem)의 구조에 적용할 때, 인공 지능 내에서 예측 모델의 불확실성을 관리하는 정교한 접근 방식이 밝혀집니다.\\
\midrule
Llama-3-Open-Ko-8B & 신경 과제 기반 모델링(neural task-driven modeling)이  \textcolor{blue}{데 핀에티의 정리}(de finetti's theorem)의 구성 요소에 적용될 때, 인공 지능 내 예측 모델의 불확실성을 관리하는 정교한 접근 방식을 드러냅니다.  \\
\midrule
Llama-3-KoEn-Instruct & 신경 작업 주도 모델링(neural task-driven modeling)을 \textcolor{blue}{데 피니티의 정리}(de finetti's theorem)의 구성에 적용할 때, 인공 지능 내에서 예측 모델의 불확실성을 관리하는 정교한 접근 방식을 드러냅니다.  \\
\bottomrule
\end{tabular}
\caption{Model output comparison for in-domain test data across various models, with sLMs fine-tuned using QLoRA. Blue text highlights the translations of specific terms that models frequently translate inconsistently, reflecting challenges in handling proper nouns.}
\label{tab:result_proper_nouns}
\end{table*}

\begin{table*}[]
\centering
\fontsize{8.5pt}{13pt}\selectfont 
\begin{tabular}{l|p{13cm}}
\toprule
\textbf{Domain} & High-Energy Physics (out-of-domain) \\
\midrule
\textbf{Terms} & jet quenching, quarkonia suppression\\
\midrule
\textbf{Source} & The interplay between jet quenching and quarkonia suppression offers a comprehensive understanding of the dynamics governing the quark-gluon plasma and its evolution. \\
\midrule
\textbf{Target} & \textcolor{Green}{제트 퀀칭}(jet quenching)과 \textcolor{Green}{쿼크오니아 억제}(quarkonia suppression) 간의 상호작용은 쿼크-글루온 플라즈마의 동역학과 그 진화에 대한 포괄적인 이해를 제공합니다.  \\
\toprule
\textbf{Model} & \textbf{Model Output} \\
\midrule
mBART50& \textcolor{red}{ject quenching}(jet quenching)과 \textcolor{red}{quarkonia suppression}(quarkonia suppression) 간의 상호 작용은 쿼크- 글루언 플라즈마와 그 진화에 대한 종합적인 이해를 제공합니다.  \\
\midrule
M2M100 & \textcolor{red}{jet quenching}과 \textcolor{red}{quarkonia suppression} 간의 상호 작용은 \textcolor{red}{quark-gluon plasma}의 역학과 그 진화에 대한 포괄적인 이해를 제공합니다. \\
\midrule
NLLB-200 & \textcolor{red}{제트 화}와 \textcolor{red}{쿼크니아} 억제 간의 상호 작용은 쿼크-글루온 플라즈마와 그 진화에 대한 동적에 대한 포괄적인 이해를 제공합니다.  \\
\midrule
\midrule
Llama 3-8B & 제트 \textcolor{red}{.jet 쿼ンチ잉}(jet quenching)과 쿼크오니아 억제(quarkonia suppression) 간의 상호 작용은 쿼크-글루온 플라즈마와 그 진화가 지배하는 역학에 대한 종합적인 이해를 제공합니다.  \\
\midrule
Gemma2-2B & 제트 격리(jet quenching)와 쿼크니아 억제(quarkonia suppression) 간의 상호 작용은 쿼크-글루온 플라즈마와 그 변화에 관한 역학을 종합적으로 이해하는 데 도움을 줍니다. \\
\midrule
Gemma2-9B & 제트 \textcolor{red}{콸닝}(jet quenching)과 쿼크니아 억제(quarkonia suppression) 간의 상호 작용은 쿼크-글루온 플라즈마와 그 진화를 지배하는 역학에 대한 포괄적인 이해를 제공합니다.  \\
\midrule
Gemma2-27B & 제트 냉각(jet quenching)과 쿼크오니아 억제(quarkonia suppression) 간의 상호 작용은 쿼크-글루온 플라즈마와 그 진화를 지배하는 역학에 대한 종합적인 이해를 제공합니다. \\
\midrule
Qwen2-1.5B &  \textcolor{red}{\_jet quenching}와\textcolor{red}{\_quetsquon suppression} 간의 상호작용은 광자론 합성(quarkonia suppression)의 동적을 종합적으로 이해하는 데 도움을 줍니다. \\
\midrule
Qwen2-7B & \textcolor{red}{점프 퀀터링}(jet quenching)과 \textcolor{red}{퀀코나임}(quarkonia suppression) 간의 상호 작용은 퀀크-글루온 플라즈마와 그 진화를 지배하는 역학을 포괄적으로 이해하는 데 중요한 역할을 합니다.  \\
\midrule
Qwen2-72B & 제트 \textcolor{red}{쿠enching}(jet quenching)과 쿼크니아 억제(quarkonia suppression) 간의 상호 작용은 쿼크-글루온 플라스마와 그 진화를 지배하는 역학에 대한 종합적인 이해를 제공합니다. \\
\bottomrule
\end{tabular}
\caption{Model output comparison for out-of-domain data across various models, with sLMs fine-tuned using LoRA. Red text highlights specific terms that are frequently mistranslated, indicating challenges in handling these out-of-domain terms.}
\label{tab:deatiled_result_ood}
\end{table*}

\begin{table*}[h!]
\centering
\begin{tabular}{lcccccc}
\toprule
\textbf{Model} & \textbf{\# Params} & \textbf{Training Technique} & \textbf{BLEU} & \textbf{COMET} & \textbf{BERT} \\

\midrule
mBART50& 611M & Full Fine-Tuning & 40.298 & 0.892 & 0.927 \\
M2M100 & 418M & Full Fine-Tuning & 41.789 & 0.892 & 0.928 \\
NLLB-200 & 600M & Full Fine-Tuning & 35.843 & 0.886 & 0.920 \\
\midrule
Llama3 & 8B & LoRA & 39.243 & 0.892 & 0.924 \\
& 8B & QLoRA & 40.984 & 0.893 & 0.927 \\
& 70B & LoRA & 40.600 & 0.893 & 0.928 \\
Llama3-Instruct & 8B-it & QLoRA & 39.686 & 0.892 & 0.924 \\
& 8B-it & 1-shot prompting & 1.103 & 0.410 & 0.594 \\
\midrule
Gemma2 & 2B & LoRA & 40.126 & 0.890 & 0.925 \\
& 9B & LoRA & 43.391 & 0.896 & 0.932 \\
& 9B & QLoRA & 41.620 & 0.893 & 0.929 \\
& 27B & LoRA & 42.313 & 0.896 & 0.931 \\
Gemma2-Instruct & 9B-it & QLoRA & 41.143 & 0.891 & 0.928 \\
& 9B-it & 1-shot prompting & 28.314 & 0.837 & 0.838 \\
\midrule
Qwen2 & 1.5B & LoRA & 36.174 & 0.881 & 0.914 \\
& 7B & LoRA & 41.434 & 0.893 & 0.927 \\
& 7B & QLoRA & 39.975 & 0.890 & 0.924 \\
& 72B & LoRA & 42.704 & 0.894 & 0.929 \\
Qwen2-Instruct & 7B-it & QLoRA & 40.107 & 0.889 & 0.923 \\
\midrule
Mistral & 7B & QLoRA & 40.424 & 0.891 & 0.925 \\
Mistral-Instruct & 7B-it & QLoRA & 39.368 & 0.891 & 0.924 \\
\midrule
Llama-3-KO-EN & 8B-it & QLoRA & 42.862 & 0.896 & 0.931 \\
& 8B-it & 1-shot prompting & 2.031 & 0.490 & 0.673 \\
Llama-3-Open\_Ko & 8B-it & QLoRA & 41.793 & 0.894 & 0.928 \\
\midrule
GPT-4o & Unknown & 0-shot prompting & 33.406 & 0.889 & 0.915 \\
GPT-4o & Unknown & 1-shot prompting & 35.272 & 0.890 & 0.918 \\
\bottomrule
\end{tabular}
\caption{Pure translation metrics \( M\) for in-domain test data. The suffix '-it' indicates instruct-tuned models.}

\label{tab:raw_performance}
\end{table*}

\end{document}